\documentclass{article}

\usepackage{arxiv}

\usepackage[utf8]{inputenc} 
\usepackage[T1]{fontenc}    
\usepackage{hyperref}       
\usepackage{url}            
\usepackage{booktabs}       
\usepackage{amsfonts}       
\usepackage{nicefrac}       
\usepackage{microtype}      
\usepackage{lipsum}
\usepackage{graphicx}
\usepackage{tabularx}
\usepackage{amsmath}
\usepackage{footmisc}
\usepackage{doi}
\usepackage{multirow}
\graphicspath{ {./images/} }

\title{Structured Decomposition for LLM Reasoning: Cross-Domain Validation and Semantic Web Integration}

\author{
 Albert Sadowski \\
  Warsaw University of Technology \\
  Warsaw, Poland \\
  \texttt{albert.sadowski.stud@pw.edu.pl} \\
   \And
 Jarosław A. Chudziak \\
  Warsaw University of Technology \\
  Warsaw, Poland \\
  \texttt{jaroslaw.chudziak@pw.edu.pl} \\
}

\begin{document}
\maketitle
\begin{abstract}
Rule-based reasoning over natural language input arises in domains where decisions must be auditable and justifiable: clinical protocols specify eligibility criteria in prose, evidence rules define admissibility through textual conditions, and scientific standards dictate methodological requirements. Applying rules to such inputs demands both interpretive flexibility and formal guarantees. Large language models (LLMs) provide flexibility but cannot ensure consistent rule application; symbolic systems provide guarantees but require structured input. This paper presents an integration pattern that combines these strengths: LLMs serve as ontology population engines, translating unstructured text into ABox assertions according to expert-authored TBox specifications, while SWRL-based reasoners apply rules with deterministic guarantees. The framework decomposes reasoning into entity identification, assertion extraction, and symbolic verification, with task definitions grounded in OWL~2 ontologies. Experiments across three domains (legal hearsay determination, scientific method-task application, clinical trial eligibility) and eleven language models validate the approach. Structured decomposition achieves statistically significant improvements over few-shot prompting in aggregate, with gains observed across all three domains. An ablation study confirms that symbolic verification provides substantial benefit beyond structured prompting alone. The populated ABox integrates with standard semantic web tooling for inspection and querying, positioning the framework for richer inference patterns that simpler formalisms cannot express.
\end{abstract}

\keywords{Large Language Models \and Explainable AI \and Neural-Symbolic Integration \and Semantic Web}

\section{Introduction}
\label{section:introduction}

Artificial intelligence systems are increasingly deployed to support decision-making in domains governed by explicit rules: clinical protocols specify patient eligibility criteria, evidence law defines admissibility conditions, and scientific standards dictate methodological requirements \cite{Lai2023HumanAIDecisionMaking}. These rule frameworks exist precisely because decisions in such domains must be auditable and justifiable~\cite{Dahl2024}. The growing volume of unstructured text, medical records, legal filings, research literature, creates demand for systems that can apply domain rules to natural language inputs at scale.

This demand exposes a fundamental tension between two computational paradigms~\cite{Wei2025, Lalwani2024}. Large language models offer unprecedented interpretive flexibility: they can identify entities, extract relationships, and resolve ambiguities inherent in natural language~\cite{Wei2022}. However, they cannot provide formal guarantees, their reasoning is opaque, their rule application inconsistent across phrasings~\cite{Mu2023, GSMSymbolic}, and their outputs unverifiable against logical specifications~\cite{Bubeck2023}. The persistence of hallucination, even in state-of-the-art models, compounds these concerns~\cite{Magesh2024}. Symbolic systems offer the inverse profile: formal guarantees through explicit inference rules, but an inability to interpret unstructured text directly~\cite{Calanzone2024, BenchCapon1997}.

The capabilities of modern LLMs have renewed interest in how symbolic reasoning systems can be leveraged~\cite{Tan2024, Kostka2024}. Rather than viewing neural and symbolic approaches as competing paradigms, recent work explores architectures that exploit their complementary strengths~\cite{Noguer2024, Servantez2024}. The critical insight is that LLMs excel at the interpretive work symbolic systems cannot perform, while symbolic systems provide the verification guarantees LLMs cannot offer. This complementarity suggests an integration strategy: use LLMs to translate unstructured text into formal representations, then delegate reasoning to symbolic systems.

We argue that OWL 2 ontologies~\cite{OWL2Primer} with SWRL rules~\cite{SWRL2004} provide the right abstraction for this integration. The architecture separates concerns cleanly: expert-authored TBox specifications encode domain rules and classification criteria; LLMs populate the ABox by extracting entities and assertions from text; SWRL-based reasoners apply rules with deterministic guarantees. This division positions LLMs as ontology population engines, components that translate unstructured input into formal assertions according to predefined specifications. The populated ABox integrates with standard semantic web tooling for inspection, querying, and further inference, addressing the interoperability limitations of ad-hoc formalisms used in prior neural-symbolic approaches.

Our prior work~\cite{sadowski2025explainable} introduced structured decomposition for rule-based reasoning, separating the process into entity identification, assertion extraction, and symbolic verification. That work demonstrated improvements on legal hearsay determination but was limited to a single task and relied on ad-hoc representations with SMT-based verification. This paper extends the framework in two directions. First, we ground task definitions in OWL 2 ontologies with SWRL rules, replacing ad-hoc representations with semantic web standards. This reformulation enables integration with established tooling and positions the framework for richer inference patterns: class hierarchies, multi-label classification, and composable imports from domain vocabularies. Second, we validate the approach across three domains, legal reasoning, scientific literature analysis, and medical inference, evaluating eleven language models to assess cross-domain generalisability.

The paper makes two primary contributions. We present a neural-symbolic integration pattern grounded in semantic web standards, demonstrating that LLMs can serve as ontology population engines for rule-based reasoning tasks. The architecture produces inspectable reasoning traces where every extraction decision persists as an ABox assertion, enabling systematic review of how text was interpreted. We then validate this pattern across three domains: legal reasoning (hearsay determination), scientific literature analysis (method-task application), and medical inference (clinical trial eligibility). Experiments across eleven language models confirm that structured decomposition generalises beyond legal reasoning, with improvements across all three domains and aggregate gains reaching statistical significance.

The remainder of this paper is organised as follows. Section~\ref{section:background} reviews related work on LLM reasoning, prompting strategies, and neural-symbolic integration, then introduces the semantic web technologies grounding our reformulation. Section~\ref{section:framework} presents the structured decomposition framework. Section~\ref{sec:task-formalisation} describes task formalisation and selection criteria. Section~\ref{sec:experimental-setup} details experimental design, and Section~\ref{section:results} presents results. Section~\ref{section:discussion} discusses findings and limitations, and Section~\ref{section:conclusion} concludes.

\section{Background and Related Work}
\label{section:background}

The promise of combining neural flexibility with symbolic guarantees has motivated decades of research, but practical integration remains elusive. Large language models have shifted this landscape: their interpretive capabilities enable new forms of neural-symbolic architecture, while their well-documented reasoning limitations make external verification mechanisms newly attractive. Realising this potential requires understanding both what LLMs can and cannot do reliably, and which symbolic formalisms offer the right abstraction for integration.

\subsection{LLM Reasoning Capabilities and Limitations}

Large language models have demonstrated capabilities in complex reasoning tasks~\cite{Wei2022}, showing potential for applications requiring sophisticated text analysis and inference. However, despite these capabilities, LLMs face limitations when applications require consistent rule application~\cite{Mu2023, GSMSymbolic}, transparent reasoning processes, and verifiable outcomes~\cite{Bubeck2023}.

The fundamental challenge lies in the tension between neural and symbolic approaches to reasoning~\cite{Wei2025, Lalwani2024}. Neural methods offer flexibility and generalisation across diverse inputs but often struggle with consistency and verifiability. In contrast, symbolic systems provide formal guarantees but traditionally require structured inputs and lack adaptability to natural language~\cite{Calanzone2024}. This dichotomy becomes particularly apparent in rule application scenarios where both precise logical inference and natural language understanding are essential~\cite{BenchCapon1997}.

Three specific challenges emerge in deploying LLMs for rule-based reasoning tasks~\cite{Kant2025, Borazjanizadeh2024}. First, while LLMs can recognise patterns in training data, they lack consistent mechanisms to apply rules with logical rigour across novel situations~\cite{Mu2023}. Second, exception handling, determining when general rules do not apply due to special conditions, presents difficulties~\cite{DiSorbo2025} as it requires both rule comprehension and contextual awareness. Third, the auto-regressive generation paradigm favours local coherence over global planning~\cite{Borazjanizadeh2024}, which can lead to reasoning errors that compound across multiple inference steps.

These limitations are particularly problematic in domains requiring formal guarantees, such as regulatory compliance, clinical trial eligibility determination, and scientific information extraction, where decisions must be auditable and justifiable~\cite{Dahl2024}. The persistence of hallucination, generating plausible but incorrect outputs, remains a concern even in state-of-the-art models~\cite{Magesh2024}, motivating approaches that provide external verification mechanisms.

\subsection{Prompting Strategies for Reasoning}

To improve LLM reasoning capabilities, researchers have developed various prompting techniques. Chain-of-Thought (CoT) prompting~\cite{Wei2022} elicits intermediate reasoning steps, improving performance on tasks requiring multi-step inference, with benefits particularly pronounced in larger models.

Beyond CoT, other specialised techniques have emerged. Program-Aided Language Models (PAL)~\cite{Gao2022} leverage LLMs to generate executable code that solves reasoning problems, combining the flexibility of natural language with the precision of programming languages. ReAct~\cite{Yao2022} interleaves reasoning and acting, allowing models to interact with external environments while building coherent reasoning traces. Decomposed Prompting~\cite{DecomposedPrompting} separates complex tasks into modular sub-problems, each handled by specialised prompts.

In domain-specific applications, Servantez et al.~\cite{Servantez2024} introduced the Chain of Logic prompting method, which separates rule-based reasoning into independent logical steps and recomposes them to form coherent conclusions. Similarly, the InsurLE framework by Cummins et al.~\cite{Cummins2025} uses controlled natural language to codify insurance contracts while preserving syntactic nuances and exposing the underlying formal logic.

A distinction exists between prompting strategies that structure output format versus those that change the reasoning architecture. Techniques like CoT primarily affect how models articulate their reasoning, while approaches that integrate external verification components—such as SMT solvers or logic programming systems—introduce architectural changes that provide formal guarantees absent from prompting alone. Our structured decomposition approach falls into the latter category.

\subsection{Neural-Symbolic Integration Approaches}

Neural-symbolic approaches combine the learning capabilities of neural networks with the reasoning power of symbolic systems. This integration is promising for rule-governed domains as it can combine LLMs' natural language understanding with the formal reasoning capabilities of logic-based systems.

Tan et al.~\cite{Tan2024} enhanced LLM reasoning through a self-driven Prolog-based CoT mechanism that iteratively refines logical inferences. Wei et al.~\cite{Wei2025} proposed a hybrid neural-symbolic framework that synergises neural representations with explicit logical rules for legal reasoning in automated systems. Similar integration approaches have been explored in multi-agent contexts, where Answer Set Programming combined with graph knowledge bases enables enhanced collaborative reasoning capabilities~\cite{Kostka2024}. Subsequent work by the same authors demonstrates that augmenting such systems with Theory of Mind mechanisms and structured critique yields synergistic improvements in reasoning quality~\cite{kostka2025cognitive}.

\begin{figure*}
\setlength{\fboxsep}{0pt}%
\setlength{\fboxrule}{0pt}%
\begin{center}
\includegraphics[width=\linewidth]{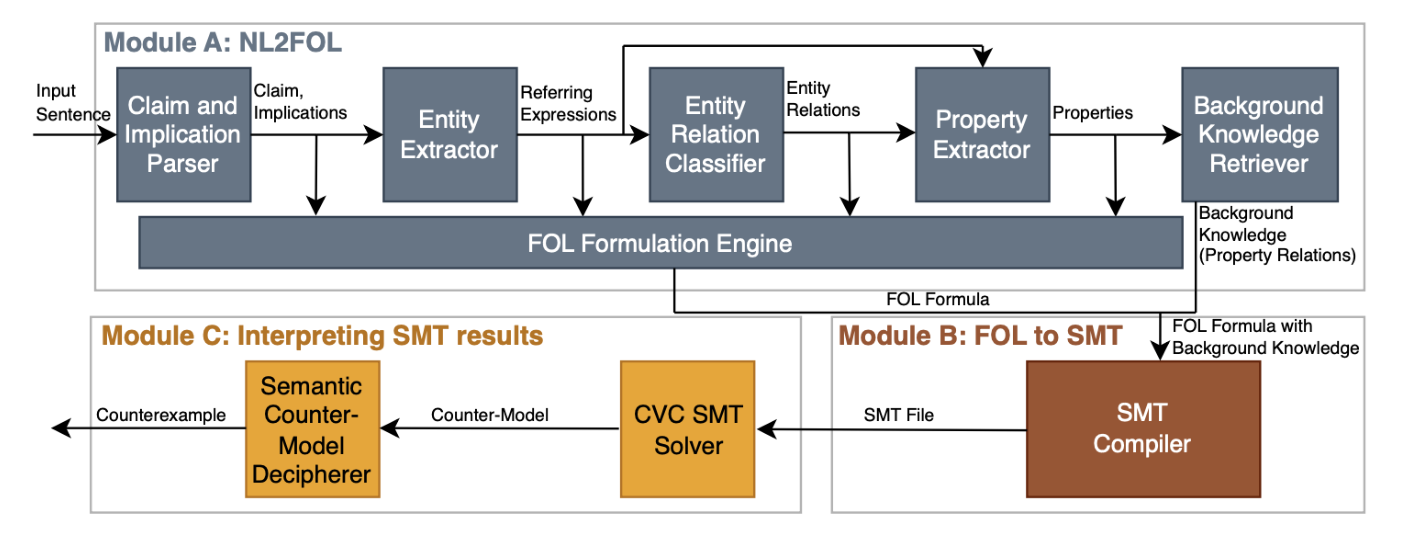}
\end{center}
\caption{The NL2FOL pipeline for translating natural language to first-order logic (reproduced from Lalwani et al.~\cite{Lalwani2024}). Module A decomposes input text through successive extraction stages, claim parsing, entity identification, relation classification, and property extraction, before formulating FOL expressions augmented with background knowledge. Module B compiles the logical formula to SMT format, and Module C interprets solver outputs as natural language counterexamples. This stepwise architecture exemplifies the neural-symbolic integration pattern our framework adopts, though we replace ad-hoc FOL translation with OWL ontology population and SMT verification with SWRL-based reasoning.}
\label{fig:nl2fol}
\end{figure*}

The work by Lalwani et al.~\cite{Lalwani2024} on NL2FOL demonstrates a structured, step-by-step pipeline to translate natural language inputs into first-order logic representations using LLMs at each step (Figure~\ref{fig:nl2fol}). Their approach addresses key challenges in this translation process, including integrating implicit background knowledge. By leveraging structured representations, they use Satisfiability Modulo Theory (SMT) solvers to reason about the logical validity of natural language statements.

Alonso and Chatzianastasiou~\cite{Noguer2024} demonstrated that embedding logical rules into neural frameworks can enhance the interpretability and robustness of text analysis. Calanzone et al.~\cite{Calanzone2024} developed an integration approach that enforces logical consistency by incorporating external constraint sets into LLM outputs.

\begin{table}[t]
\caption{Comparison of neural-symbolic methods along domain, decomposition, and symbolic reasoning dimensions.}
\label{tab:ns_comparison}
\begin{tabularx}{\hsize}{@{}lX p{4cm} p{4cm} @{}}
\toprule
\textbf{Method} & \textbf{Domain} & \textbf{Decomposition Strategy} & \textbf{Symbolic Formalism} \\
\midrule
Thought-Like-Pro~\cite{Tan2024} & General & Explicit multi-step (CoT-like) & Prolog-based inference \\
Wei et al. (2025)~\cite{Wei2025} & Legal (civil) & Structured pipeline & FOL with custom legal rules \\
NL2FOL~\cite{Lalwani2024} & General & Stepwise formalisation & FOL + SMT solver \\
Alonso \& Chatzianastasiou (2024)~\cite{Noguer2024} & Legal (contracts) & Logic-LLM integration & Logic rule templates \\
Calanzone et al. (2024)~\cite{Calanzone2024} & General & End-to-end constraint loss & Logical consistency constraints \\
Our prior work~\cite{sadowski2025explainable} & Legal (hearsay) & Three-step decomposition & SMT solver (ad-hoc schema) \\
\textbf{This work} & Cross-domain & Three-step decomposition & OWL 2 + SWRL \\
\bottomrule
\end{tabularx}
\end{table}

Table~\ref{tab:ns_comparison} provides a comparative overview of representative approaches. The existing landscape reveals a gap between systems using custom symbolic formalisms and those leveraging established semantic web standards. Most approaches employ ad-hoc representations or domain-specific logic programming systems, limiting interoperability and reuse. Our prior work~\cite{sadowski2025explainable} used JSON schemas with SMT verification, which achieved strong results but lacked standardised semantics and tooling support. This observation motivates grounding structured decomposition in OWL and SWRL, enabling integration with the broader semantic web ecosystem.

\subsection{Semantic Web Technologies for Knowledge Representation}

The semantic web provides a mature technology stack for formal knowledge representation and reasoning. While originally designed for web-scale interoperability, these standards have increasingly found application in neuro-symbolic architectures as the symbolic substrate for verification and logic injection~\cite{HitzlerSarker2022}. We introduce the core technologies relevant to our framework reformulation.

OWL 2~\cite{OWL2Primer} is a W3C standard for defining ontologies, formal specifications of shared conceptualisations within a domain. OWL ontologies consist of two components: the TBox (terminological box) defines classes, properties, and their relationships; the ABox (assertional box) contains individuals and their property assertions. This separation, rooted in Description Logic theory~\cite{Baader2007}, maps naturally to our framework's distinction between task definitions (TBox) and extracted entities/predicates (ABox).

OWL provides constructs for defining class hierarchies, property domains and ranges, and logical constraints. For example, defining \textit{Hearsay} as a subclass of \textit{Statement} captures the taxonomic relationship that hearsay evidence is a specialised form of statement. Property restrictions can express necessary conditions for class membership, though complex rule-based inference requires additional machinery.

SWRL~\cite{SWRL2004} extends OWL with Horn-like rules, enabling inference patterns beyond OWL's native expressivity. A SWRL rule takes the form: \textit{antecedent $\rightarrow$ consequent}, where both antecedent and consequent are conjunctions of atoms. Atoms can be OWL class assertions, property assertions, or built-in predicates for comparisons and computations.

For structured decomposition, SWRL rules encode the task predicate, the logical formula determining when a classification holds. For instance, the hearsay determination rule can be expressed as:

\begin{equation}
\small
\begin{split}
&\texttt{Statement}(?s) \land \texttt{OutOfCourtStatement}(?s) \\
&\quad \land\; \texttt{hasAssertion}(?s, ?a) \\
&\quad \land\; \texttt{introducedForLegalIssue}(?s, ?l) \\
&\quad \land\; \texttt{provesTruthOfAssertion}(?s, ?l) \\
&\quad \rightarrow \texttt{Hearsay}(?s)
\end{split}
\label{eq:hearsay-fol}
\end{equation}

This rule states that if an individual \textit{?s} satisfies all antecedent conditions, it is inferred to belong to the \textit{Hearsay} class.

OWL reasoning operates under the open-world assumption (OWA): the absence of a statement does not imply its negation. This contrasts with databases and many logic programming systems that use the closed-world assumption (CWA). Under OWA, if a property is not asserted for an individual, the reasoner cannot conclude that the property does not hold; it may simply be unknown~\cite{Hitzler2009}.

For the binary classification tasks in this paper, the practical impact is limited: SWRL rules with conjunctive antecedents require all conditions to be satisfied for the rule to fire. Whether a missing assertion is treated as ``unknown'' (OWA) or ``false'' (CWA), the rule does not fire in either case, yielding the same classification outcome. More complex inference patterns, such as rules that reason over the absence of properties, would require explicit handling of the OWA.

Several OWL reasoners are available, including Pellet~\cite{Sirin2007}, HermiT~\cite{Glimm2014}, and FaCT++~\cite{Tsarkov2006}. These reasoners support consistency checking, classification, and SWRL rule execution. We use the Pellet reasoner, which integrates with the owlready2 library used in our implementation.

\subsection{LLMs for Ontology Engineering and Population}

Recent work has explored using LLMs for ontology-related tasks, reflecting a broader shift in the semantic web community toward leveraging large language models for tasks traditionally requiring significant manual engineering~\cite{Pan2024}.

\subsubsection{Ontology Learning and Alignment}

LLMs have been applied to ontology learning, the automatic or semi-automatic construction of ontologies from text. The \textit{LLMs4OL} paradigm~\cite{Babaei2023} demonstrated that LLMs can extract taxonomic relations, identify relevant concepts, and propose property definitions from unstructured corpora with competitive performance. Recent benchmark studies confirm that while models excel at surface-level extraction, correctly identifying complex axioms remains challenging without structured guidance~\cite{Bakker2025}. Consequently, resulting ontologies often require expert refinement to ensure logical consistency and domain accuracy.

Ontology alignment, mapping concepts between different ontologies, has similarly benefited from LLM capabilities. Models can identify semantic correspondences based on contextual understanding rather than purely lexical matching~\cite{Hertling2023}. These capabilities position LLMs as useful tools in the ontology engineering process, though recent position papers argue they should not replace formal verification in high-stakes contexts~\cite{Herron2025}.

\subsubsection{Knowledge Graph Population}

Knowledge graph population involves extracting entities and relations from text to populate a knowledge base, a task closely aligned with our framework's entity identification and assertion extraction steps. Studies have shown LLMs can achieve competitive performance on relation extraction benchmarks~\cite{Wadhwa2023}.

Despite progress in these individual areas, systematic frameworks that integrate LLMs as ontology population components with formal OWL/SWRL verification remain underexplored. Most existing work either uses LLMs for ontology construction (producing TBox definitions) or for standalone information extraction (without formal verification). Approaches like NL2FOL (Figure~\ref{fig:nl2fol}) employ ad-hoc FOL representations and custom SMT compilation, limiting interoperability with established semantic web infrastructure.

\section{Structured Decomposition Framework}
\label{section:framework}

The integration pattern rests on a clean separation of concerns: LLMs handle interpretation, symbolic reasoners handle inference, and ontologies provide the interface between them. Our prior work~\cite{sadowski2025explainable} demonstrated this separation using ad-hoc JSON schemas with SMT verification. Here we reformulate the framework around OWL~2 ontologies and SWRL rules, replacing custom representations with semantic web standards. This grounds the architecture in established tooling while preserving its reasoning guarantees.

\subsection{Design Principles and Architecture Overview}

Decomposition-based approaches have long been recognised in cognitive science and AI as effective strategies for complex problem solving~\cite{Newell1972, Sacerdoti1974}. Our framework applies these principles to address a fundamental challenge: while LLMs demonstrate impressive capabilities with natural language inputs, they struggle with consistent rule application~\cite{GSMSymbolic, Mu2023}.

The framework rests on three design principles. First, \textit{separation of concerns} divides the reasoning pipeline between neural and symbolic components. LLMs handle language interpretation, recognising entities and evaluating assertions from unstructured text, while ontological reasoners handle logical inference with formal guarantees. Each component does what it does best. Second, \textit{externalised task definitions} allow domain experts to specify terms, predicates, and classification rules without modifying the framework architecture. Experts can refine definitions as interpretations evolve, accommodating the ``open texture'' inherent in many domain concepts~\cite{BenchCapon1997}. Third, \textit{inspectable intermediate state} arises naturally from the step-wise decomposition. Each extraction step populates the ABox with explicit assertions, identified entities, extracted assertions, inferred classifications, that persist as a reviewable record. Unlike end-to-end approaches where reasoning is opaque, here every interpretive decision is externalised and available for scrutiny.

\begin{figure*}
\setlength{\fboxsep}{0pt}%
\setlength{\fboxrule}{0pt}%
\begin{center}
\includegraphics[width=\linewidth]{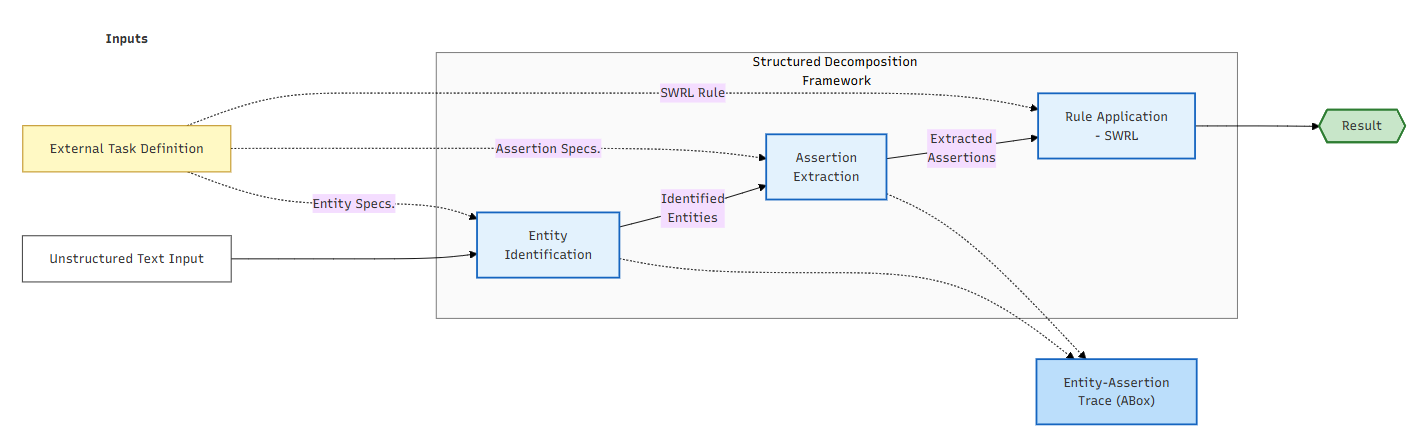}
\end{center}
\caption{Structured Decomposition Framework with OWL/SWRL integration. Yellow box represents domain expert input (task ontology defining classes, properties, and SWRL rules); blue boxes show automated processing by LLMs (entity identification, assertion extraction) and reasoner (rule application); arrows indicate data flow. Beyond the classification outcome, the framework yields a populated ABox that externalises every extraction decision, providing an inspectable record of how the input text was interpreted.}
\label{fig:structured_decomposition_framework}
\end{figure*}

Figure~\ref{fig:structured_decomposition_framework} illustrates the architecture. The pipeline comprises three steps: entity identification extracts individuals from text according to class definitions; assertion extraction determines which properties hold for those individuals; rule application uses an OWL reasoner to infer classifications based on SWRL rules encoding the task predicate. The task definition, formalised as an OWL ontology, provides the structural scaffold for all three steps. This decomposition mirrors the modular design of prior neural-symbolic pipelines (cf.\ Figure~\ref{fig:nl2fol}), but grounds extraction in ontology-defined specifications and replaces SMT verification with SWRL-based reasoning.

This architecture positions LLMs as ontology population engines, components that translate unstructured text into ABox assertions according to TBox definitions provided by the task ontology. The reasoner handles inference over the populated ontology, applying SWRL rules to determine final classifications. This division exploits the complementary strengths of each component while localizing the sources of potential error.

\subsection{Entity Identification Step}

The entity identification step extracts individuals from unstructured text corresponding to classes defined in the task ontology. Given input text and entity specifications from the ontology, the LLM identifies text spans matching each entity type, assigns extracted entities to appropriate classes, and provides explanations justifying each extraction.

The prompt structure supplies the LLM with entity specifications expressed in natural language. Each entity type is defined with properties specifying the entity type name, a natural language description guiding extraction, and a mapping to the corresponding OWL class. For example, an entity specification might specify: ``A communicative act (verbal, written, or non-verbal conduct) intended to assert a fact. Detectable by explicit claims (e.g., `Alex said...'), written documents, or purposeful actions (e.g., nodding).'' These descriptions provide the LLM with both conceptual definitions and practical detection heuristics.

Entity identification maps directly to ABox population: each extracted entity becomes an individual in the ontology, with class membership assertions reflecting the LLM's determinations. The step produces structured output specifying, for each defined entity type, the extracted text span (if any), the assigned individual identifier, and a natural language explanation grounding the extraction in the input text.

Several challenges arise in this step. Entity boundary detection requires determining where relevant text spans begin and end, particularly for complex or embedded references. Implicit entities, those inferable from context but not explicitly stated, demand interpretive judgment from the LLM. The explanation requirement serves both transparency and quality assurance: explanations that fail to coherently justify extractions signal potential errors for human review.

\subsection{Assertion Extraction Step}
\label{subsec:assertion-extraction}

The assertion extraction step determines which properties and relationships hold between identified entities. Given the entities from the previous step and assertion specifications from the task ontology, the LLM evaluates each specification, produces binary determinations (true/false), and provides justifications grounding each decision in textual evidence.

Assertion specifications define the assertion name, the entity types it relates, a natural language description, and a mapping to either an OWL class (for unary assertions representing class membership) or an OWL object property (for binary assertions representing relationships). The description provides guidance for evaluation; for instance, an assertion specification determining whether a statement occurred out of court might specify: ``True if the statement was made outside the current trial/hearing. Detect via contextual cues (e.g., ``told his brother'' vs. ``testified in court'').''

A design pattern we investigate is the use of \textit{complementary predicates}, paired assertions representing opposing determinations. This pattern addresses a practical challenge arising from OWL's open-world assumption (OWA): absent an assertion, the reasoner cannot conclude that a property does not hold, it may simply be unknown. Complementary predicates force the LLM to make explicit determinations on both positive and negative conditions, converting implicit assumptions into explicit assertions.

In our prior work~\cite{sadowski2025explainable}, complementary predicates yielded substantial precision improvements on hearsay determination by mitigating confirmation bias. This approach reduced false positives by forcing the explicit consideration of both confirming and disconfirming factors. As it remains unclear whether this benefit extends to other domains or persists with more capable model architectures, we treat complementary predicates as an experimental variable rather than a default component, evaluating the framework both with and without them across all three tasks.

Assertion extraction produces structured output specifying, for each assertion specification, the entities involved, the boolean determination, and a natural language justification citing specific textual evidence. These outputs map to class membership or property assertions in the ontology's ABox.

\subsection{OWL Ontology Formulation}

Task definitions are formalised as OWL~2 ontologies, replacing the ad-hoc JSON schemas used in our prior work. Task specifications become editable in tools such as Protégé, and populated ABoxes conform to standards enabling integration with domain ontologies and SPARQL-based querying.

Each task ontology comprises a TBox defining the conceptual structure and an ABox populated during inference. The TBox includes class hierarchies defining entity types relevant to the task, with the target classification modeled as a class (typically a subclass of a more general entity class). Object properties capture relationships between entities, with domain and range restrictions constraining valid assertions.

Classification logic is encoded as SWRL rules. Each rule specifies when an individual belongs to the target class: the antecedent is a conjunction of class membership tests and property assertions; the consequent asserts membership in the target class. For example, the hearsay rule encodes the Federal Rules of Evidence definition, when all conditions are satisfied, the individual is classified as hearsay.

Task-level metadata for LLM prompting (entity descriptions, assertion specifications, and domain context) are encoded as annotations within the ontology, separating prompt generation concerns from OWL reasoning semantics. This separation allows the same ontology to drive both symbolic verification and neural interpretation. Domain experts can edit task definitions using standard ontology tooling, validate logical consistency, and refine natural language descriptions iteratively without modifying the framework implementation. The complete ontologies for all three evaluation tasks are provided in the supplementary materials.\footnote{\label{fn:sd}\url{https://github.com/albsadowski/structured-decomposition-swj}}

\subsection{Reasoning and ABox Inspection}

The rule application step invokes an OWL reasoner to evaluate SWRL rules over the populated ontology. We use Pellet~\cite{Sirin2007} for its SWRL support and integration with standard ontology APIs, though other reasoners supporting SWRL could substitute.

Given the ABox assertions from entity identification and assertion extraction, the reasoner performs consistency checking, classification, and SWRL rule execution. The classification outcome, whether the target individual belongs to the target class, constitutes the framework's prediction.

The populated ABox serves as an inspectable artifact grounding each classification decision. Every entity extracted and every assertion evaluated persists as an explicit record. SPARQL queries can retrieve these assertions systematically; for instance, after processing the hearsay evaluation dataset:

\begin{verbatim}
PREFIX : <http://example.org/hearsay#>
SELECT ?case ?statement ?assertion
WHERE {
  ?statement :belongsToCase ?case ;
             a :OutOfCourtStatement ;
             :hasAssertion ?assertion .
}
\end{verbatim}

This query retrieves out-of-court statements along with their asserted content, returning instances such as ``told his brother that...,'' ``wrote in the letter...,'' enabling systematic review of how the input text was interpreted. The natural language justifications generated during extraction provide human-readable rationales alongside these formal records. For regulated domains requiring audit trails, the framework produces a complete record of interpretive decisions rather than a black-box prediction.

Because the symbolic reasoner always produces the same output for a given set of assertions, any variability in the framework's behaviour stems solely from the LLM-based extraction steps. This separation makes the source of uncertainty easy to locate: reviewers can examine the extracted entities and assertions, along with their natural language justifications, to identify where interpretive judgments were made.

\section{Task Formalisation}
\label{sec:task-formalisation}

Structured decomposition targets a specific class of problems: those where authoritative rules fully determine classification and where the relevant predicates can be extracted from text. We formalise three such tasks spanning legal, scientific, and medical domains. Each satisfies both criteria, enabling evaluation of whether framework benefits generalise beyond any single domain. The complete ontologies are available in the supplementary materials.\footref{fn:sd}

\subsection{Task Selection and Characterization}
\label{subsec:task-selection}

Not all classification problems are amenable to structured decomposition. The approach requires specific properties that distinguish rule-governed reasoning from statistical pattern matching. Our analysis identifies two necessary conditions.

First, the task must have a \emph{rule-expressible decision boundary}: the classification must be fully determined by a logical formula over extractable predicates. Rules must not only exist but be sufficient to determine the outcome. Second, the domain must permit \emph{formalisable predicate structure}, allowing decomposition into discrete predicates whose logical composition captures necessary and sufficient conditions for classification. These conditions are conjunctive: violating either renders structured decomposition inappropriate.

Table~\ref{tab:task-suitability} summarizes task suitability across four candidate domains. The three tasks satisfying both criteria form our evaluation set: hearsay determination from legal reasoning, method application identification from scientific text, and clinical trial eligibility inference from the medical domain. Each involves authoritative rule sources (Federal Rules of Evidence, relation annotation guidelines, clinical trial protocols) that provide deterministic classification criteria.

\begin{table}[htbp]
\centering
\caption{Task suitability for structured decomposition. Tasks require both a rule-expressible decision boundary and formalisable predicates.}
\label{tab:task-suitability}
\small
\begin{tabular}{@{}lcccc@{}}
\toprule
& \rotatebox{90}{\textbf{Hearsay}} & \rotatebox{90}{\textbf{Method Application}} & \rotatebox{90}{\textbf{Clinical Trial Eligibility}} & \rotatebox{90}{\textbf{URTI}} \\
\midrule
Rule-expressible boundary & \checkmark & \checkmark & \checkmark & \texttimes \\
Formalisable predicates & \checkmark & \checkmark & \checkmark & \checkmark \\
\bottomrule
\end{tabular}
\end{table}

To validate these criteria empirically, we evaluated structured decomposition on Upper Respiratory Tract Infection (URTI) classification derived from DDXPlus~\cite{fansi2022ddxplus}, a medical diagnosis task with formalisable predicates (symptoms, patient history) but no rule-expressible boundary. The results confirmed our characterization: structured decomposition achieved 0.145 F1 compared to 0.979 for few-shot prompting. The framework correctly extracted predicates according to diagnostic criteria, but those criteria do not constitute a sufficient decision boundary. When symptom profiles are consistent with multiple diagnoses, the true boundary is statistical, depending on distributional patterns that few-shot learning captures but logical rules cannot express. This negative result validates the suitability criteria: structured decomposition is a targeted approach for rule-governed domains, not a general-purpose improvement.

\subsection{Hearsay Determination}
\label{subsec:hearsay-task}

The hearsay determination task, evaluated in our prior work~\cite{sadowski2025explainable}, requires classifying whether a piece of evidence constitutes inadmissible hearsay under Rule 801 of the Federal Rules of Evidence. The rule defines hearsay as an out-of-court statement offered to prove the truth of the matter asserted. We retain the original logical formalisation:

\begin{equation}
\small
\begin{split}
\texttt{IsHearsay}(s, l) \Leftrightarrow\; & \texttt{IsStatement}(s) \\
& \quad \land \texttt{IsOutOfCourt}(s) \\
& \land\; \exists a.\big(\texttt{HasAssertion}(s, a) \\
& \quad \land \texttt{IntroducedFor}(s, l) \\
& \quad \land \texttt{ProvesTruth}(s, l)\big)
\end{split}
\label{eq:hearsay-fol}
\end{equation}

\noindent where $s$ denotes a communicative act (verbal, written, or non-verbal conduct), $l$ the disputed legal issue, and $a$ the factual claim conveyed by the statement.

The ontology defines three OWL classes: \textit{Statement}, \textit{OutOfCourtStatement}, and \textit{Hearsay}, with the latter two modeled as subclasses of \textit{Statement}. Object properties \textit{hasAssertion}, \textit{introducedForLegalIssue}, and \textit{provesTruthOfAssertion} capture the relational predicates. The SWRL rule implements Formula~\ref{eq:hearsay-fol}: if an individual satisfies all antecedent conditions, it is inferred to belong to \textit{Hearsay}.

The task uses the LegalBench hearsay dataset~\cite{LegalBench}. Consider two contrasting examples. ``To prove that Tom was in town, a witness testifies that her friend Susan told her Tom was in town'' is hearsay: an out-of-court statement is introduced to prove its truth. Conversely, ``To prove that Arthur knew English, the fact that Arthur told Bill (in English) that he thought Mary robbed the bank'' is not hearsay: the statement is introduced to prove Arthur's English proficiency, not Mary's guilt.

\subsection{Method Application}
\label{subsec:method-application-task}

The method application task evaluates whether a scientific text describes a computational method being applied to a specific task. We derive this benchmark from the SciERC dataset~\cite{luan2018scierc}, which annotates scientific abstracts with entity and relation labels. We filter to instances involving the \textit{Used-For} relation and formulate the problem as binary classification: given a text with a marked method entity and a marked task entity, determine whether the method is functionally applied to the task.

The challenge lies in distinguishing genuine application from superficial co-occurrence. Scientific abstracts frequently mention methods and tasks in conjunction (``X and Y''), comparison (``X outperforms Y''), or taxonomic contexts (``X is a type of Y'') without implying functional application. The logical formalisation captures this distinction:

\begin{equation}
\small
\begin{split}
\texttt{Applicable}(m) \Leftrightarrow\; & \texttt{IsMethod}(m) \land \texttt{IsTask}(t) \\
& \land\; \texttt{FunctionalConn}(m, t) \\
& \land\; \texttt{NotExclRelation}(m, t)
\end{split}
\label{eq:method-app-fol}
\end{equation}

\noindent where $m$ denotes the candidate method and $t$ the candidate task. The \textit{FunctionalConn} predicate is satisfied by usage verbs (``applied to'', ``enables''), purposive markers (``for the task of''), or implicit purposive constructions linking method to goal. The \textit{NotExcludedRelation} predicate filters out conjunction, comparison, part-of, hyponym, and feature-of relationships.

The ontology defines \textit{Method} and \textit{ScientificTask} as disjoint classes, with \textit{MethodApplication} modeled as a subclass of \textit{Method}. Object properties \textit{functionallyConnectsTo} and \textit{hasValidRelationTypeWith} encode the two conjuncts. The SWRL rule infers membership in \textit{MethodApplication} when both properties are satisfied.

Input instances mark entities using bracket notation inherited from dataset preprocessing: methods in double brackets (\textit{[[\ ]]}) and tasks in angle brackets (\textit{\textless\textless\ \textgreater\textgreater}). Consider two examples. ``We present an application of [[ ambiguity packing and stochastic disambiguation techniques ]] for Lexical-Functional Grammars to the domain of \textless\textless sentence condensation \textgreater\textgreater'' is positive: the method is explicitly applied via purposive language. In contrast, ``We examine the relationship between the two \textless\textless grammatical formalisms\textgreater\textgreater: [[ Tree Adjoining Grammars ]] and Head Grammars'' is negative: the relationship is comparative, not applicative.

\subsection{Clinical Trial Eligibility}
\label{subsec:eligibility-nli-task}

The clinical trial eligibility task requires determining whether a natural language statement about patient eligibility logically follows from stated inclusion and exclusion criteria. We derive this benchmark from the NLI4CT dataset~\cite{jullien2023nli4ct}, filtering to instances where the premise contains eligibility criteria and the hypothesis makes a claim about who may or may not participate. The task is formulated as binary classification: entailment (the statement follows from the criteria) or contradiction (the statement conflicts with the criteria).

This task demands precise logical reasoning over medical terminology. Inclusion criteria specify conditions patients must satisfy; exclusion criteria specify disqualifying conditions. A statement entails when its claim is a logical consequence of these criteria, accounting for synonymy (``claustrophobia'' and ``fear of confined spaces'' are equivalent) and implicit exclusion (if inclusion requires lymph node status N0 or N1, then N2 patients are excluded). The logical formalisation is:

\begin{equation}
\small
\begin{split}
\texttt{Entails}(s, p) \Leftrightarrow\; & \texttt{IsEligibilityStatement}(s) \\
& \land\; \texttt{IsEligibilityCriteria}(p) \\
& \land\; \texttt{FollowsFromPremise}(s, p)
\end{split}
\label{eq:eligibility-fol}
\end{equation}

\noindent where $s$ denotes the hypothesis statement and $p$ the premise containing trial criteria. The \textit{FollowsFromPremise} predicate requires evaluating whether the claim in $s$ can be derived from the logical structure of $p$, rather than merely checking for textual overlap.

The ontology defines \textit{EligibilityStatement} and \textit{EligibilityCriteria} as the primary classes, with \textit{Entailment} and \textit{Contradiction} modeled as disjoint subclasses of \textit{EligibilityStatement}. The SWRL rule infers membership in \textit{Entailment} when the statement is classified as following from the premise.

Unlike hearsay determination, where reasoning concerns evidentiary purpose, clinical eligibility requires domain knowledge about medical terminology and clinical trial protocols. Consider two examples. Given the exclusion criterion ``Known claustrophobia, presence of pacemaker and/or ferromagnetic material in their body that would prohibit MRI imaging'', the statement ``Patients with irrational fear of confined spaces are not eligible for the primary trial'' entails: the model must recognise synonymy between ``claustrophobia'' and ``irrational fear of confined spaces''. Conversely, given inclusion criteria specifying organ function thresholds, the statement ``xiap gene mutation'' neither follows from nor contradicts the premise, it references absent information, yielding a contradiction under the NLI4CT annotation scheme.

\subsection{Ontology Design Patterns}
\label{subsec:ontology-patterns}

The three task ontologies share a common architectural pattern. Each ontology defines OWL classes representing domain entities (statements, methods, eligibility criteria) and object properties capturing their relationships. The target classification is modeled as class membership: an individual belongs to \textit{Hearsay}, \textit{MethodApplication}, or \textit{Entailment} when the SWRL rule's antecedent conditions are satisfied. This target class is consistently defined as a subclass of the primary entity class, reflecting that not all statements are hearsay, not all methods are applied, and not all eligibility claims entail.

As described in Section~\ref{subsec:assertion-extraction}, we include complementary predicates as an experimental condition to test whether findings from our prior work generalise. Each task ontology defines appropriate assertion pairs: \textit{FunctionalConn} with \textit{NoFunctionalConn} for method application, \textit{FollowsFromPremise} with \textit{StatementConflictsWithPremise} for eligibility, and \textit{IsOutOfCourt} with \textit{IsInCourt} for hearsay.

Practitioners formalising new domains should begin by identifying the target classification and its necessary conditions, then work backward to the entities and relationships those conditions presuppose. Entity and assertion specifications require particular care, as they constitute the natural language interface between the ontology and the LLM, vague descriptions yield inconsistent extractions. The suitability criteria from Section~\ref{subsec:task-selection} should be verified before investing in ontology development: if the task lacks deterministic semantics or authoritative rule sources, structured decomposition will not succeed.

\section{Experimental Design}
\label{sec:experimental-setup}

The framework is evaluated across three reasoning tasks spanning legal, scientific, and medical domains, each satisfying the suitability criteria established in Section~\ref{sec:task-formalisation}: a rule-expressible decision boundary and formalisable predicate structure. Eleven language models and six experimental conditions yield 33 model-task combinations, enabling assessment of cross-domain generalisability, model capability requirements, and the contribution of symbolic verification through controlled ablation. Table~\ref{tab:dataset-summary} summarises the dataset characteristics.

\begin{table*}[htbp]
\centering
\caption{Dataset summary across evaluation tasks. All tasks use stratified sampling to preserve class distributions.}
\label{tab:dataset-summary}
\begin{tabular}{llcccc}
\toprule
Task & Source & Test & Train & Positive & Negative \\
\midrule
Hearsay & LegalBench~\cite{LegalBench} & 94 & 5 & 41 & 53 \\
Method Application & SciERC~\cite{luan2018scierc} & 94 & 5 & 52 & 42 \\
Clinical Trial Eligibility & NLI4CT~\cite{jullien2023nli4ct} & 94 & 4 & 47 & 47 \\
\bottomrule
\end{tabular}
\end{table*}

\paragraph{Hearsay Determination.} We use the LegalBench hearsay determination task~\cite{LegalBench}, which requires determining whether evidence constitutes inadmissible hearsay under Rule 801 of the Federal Rules of Evidence. The dataset provides natural language descriptions of evidentiary scenarios with binary labels (\textit{Yes}/\textit{No}). We use the standard train/test split provided by LegalBench: 94 test instances and 5 training instances (used as few-shot exemplars).

\paragraph{Method Application.} We derive this task from the SciERC dataset~\cite{luan2018scierc}, which annotates scientific abstracts with entity and relation labels. We filter to instances involving the \textit{Used-For} relation and formulate the problem as binary classification: given text with marked method and task entities (using bracket notation inherited from the dataset preprocessing), determine whether the method is functionally applied to the task. Instances are sampled to yield 94 test and 5 training examples.

\paragraph{Clinical Trial Eligibility.} We filter the NLI4CT dataset~\cite{jullien2023nli4ct} to instances where the premise contains eligibility-related content, identified by keywords including ``Eligibility'', ``Inclusion'', ``Exclusion'', and ``DISEASE CHARACTERISTICS''. The task requires determining whether a natural language statement about patient eligibility logically follows from stated criteria. We apply stratified sampling to balance entailment and contradiction labels, yielding 94 test instances and 4 training instances with balanced class distribution.

All datasets are processed through a unified pipeline that standardises the input format and label encoding. Text inputs are preserved in their original form without normalisation to maintain domain-specific terminology and formatting conventions. Labels are mapped to a consistent binary format (\textit{Yes}/\textit{No}) across all tasks. The preprocessing code and dataset generation scripts are available in our supplementary repository.\footnote{\label{fn:ruleeval}\url{https://github.com/albsadowski/ruleeval-xd}}

We constrain test set sizes to 94 instances per task, determined by the smallest available dataset (hearsay). This ensures balanced evaluation across domains and enables direct comparison of framework performance without confounding from unequal sample sizes.

\subsection{Model Selection and Configuration}

We evaluate our framework across eleven language models from four providers, selected to represent diverse architectural approaches and capability levels. Table~\ref{tab:models} summarises the models and their configurations.

\begin{table*}[htbp]
\centering
\caption{Models evaluated across all experimental conditions.}
\label{tab:models}
\begin{tabular}{p{3cm}lll}
\toprule
Provider & Model & Identifier \\
\midrule
\multirow{4}{*}{OpenAI} 
  & GPT-5 Nano & \texttt{gpt-5-nano} \\
  & GPT-5 Mini & \texttt{gpt-5-mini} \\
  & GPT-5.2 & \texttt{gpt-5.2} \\
  & o3 & \texttt{o3} \\
\midrule
\multirow{2}{*}{Anthropic} 
  & Claude 4.5 Sonnet & \texttt{claude-sonnet-4-5-20250929} \\
  & Claude 4.5 Haiku & \texttt{claude-haiku-4-5-20251001} \\
\midrule
\multirow{2}{*}{Google} 
  & Gemini 2.5 Flash & \texttt{gemini-2.5-flash} \\
  & Gemini 2.5 Pro & \texttt{gemini-2.5-pro} \\
\midrule
\multirow{3}{*}{Fireworks AI} 
  & Qwen 3 & \texttt{qwen3-235b-a22b-instruct-2507} \\
  & DeepSeek v3.2 & \texttt{deepseek-v3p2} \\
  & Kimi K2 & \texttt{kimi-k2-instruct-0905} \\
\bottomrule
\end{tabular}
\end{table*}

Models were selected to provide coverage across three dimensions. First, architectural diversity: we include both standard autoregressive models and reasoning-optimised variants to assess whether explicit reasoning mechanisms interact with structured decomposition. Second, capability spectrum: models range from smaller, faster variants (GPT-5 Nano, Claude 4.5 Haiku, Gemini 2.5 Flash) to larger, more capable models (GPT-5.2, Claude 4.5 Sonnet, Gemini 2.5 Pro), enabling analysis of how model capability affects framework performance. Third, provider diversity: including models from multiple providers reduces the risk that findings are artifacts of a single training methodology or model family.

All models use deterministic settings where available (temperature 0.0) to ensure reproducibility. Responses are structured via Pydantic (v2.x) models through the LangChain framework, ensuring consistent output parsing across all models and conditions.

\subsection{Experimental Conditions}

We evaluate six experimental conditions to isolate the contributions of different framework components and assess the effect of complementary predicates. Table~\ref{tab:conditions} summarises the condition matrix.

\begin{table*}[htbp]
\centering
\caption{Experimental condition matrix. SD conditions use SWRL-based reasoning; SD-Direct conditions bypass symbolic verification.}
\label{tab:conditions}
\begin{tabular}{lcc}
\toprule
Condition & Symbolic Verification & Complementary Predicates \\
\midrule
Few-Shot (FS) & --- & --- \\
Chain-of-Thought (CoT) & --- & --- \\
SD & \checkmark & --- \\
SD-Comp & \checkmark & \checkmark \\
SD-Direct & --- & --- \\
SD-Direct-Comp & --- & \checkmark \\
\bottomrule
\end{tabular}
\end{table*}

\paragraph{Few-Shot Prompting (FS).} The baseline condition provides models with training examples from each task followed by a new test case. Models receive the task description, few-shot exemplars (input-output pairs from the training split), and the test input, then generate a direct classification. This baseline represents standard practice for LLM-based classification without explicit reasoning structure.

\paragraph{Chain-of-Thought Prompting (CoT).} Models are instructed to reason step-by-step before providing a final answer. This baseline assesses whether general reasoning elicitation provides benefits comparable to our structured approach, following the methodology established by Wei et al.~\cite{Wei2022}.

\paragraph{Structured Decomposition (SD).} The framework as described in Section~\ref{section:framework}, using standard predicate definitions without complementary predicates. Models perform entity identification and assertion extraction according to task ontology specifications, with extracted predicates evaluated by the Pellet reasoner using SWRL rules.

\paragraph{Structured Decomposition with complementary predicates (SD-Comp).} Identical to SD, but with complementary predicates included in the assertion specifications. Complementary predicates are paired assertions representing opposing determinations (e.g., both \textit{FunctionalConn} and \textit{NoFunctionalConn}), designed to force the LLM to make explicit determinations on both positive and negative conditions rather than defaulting to one option. Introduced to address the confirmation bias of LLMs.

\paragraph{Structured Decomposition Direct (SD-Direct).} An ablation condition that uses identical entity and assertion specifications as SD (without complementary predicates) but bypasses the OWL/SWRL verification step. Instead, the LLM directly determines the final classification based on extracted predicates in a single additional inference call. This ablation isolates the contribution of symbolic verification.

\paragraph{Structured Decomposition Direct with complementary predicates (SD-Direct-Comp).} Identical to SD-Direct, but with complementary predicates included.

The six conditions form a $2 \times 2$ factorial design (symbolic verification $\times$ complementary predicates) plus two baselines. This design enables us to assess: (1) whether structured decomposition with SWRL verification outperforms baselines; (2) whether symbolic verification provides value beyond structured prompting alone (SD vs. SD-Direct); and (3) whether complementary predicates improve performance, and whether this effect interacts with symbolic verification (comparing the complementary predicate effect across SD and SD-Direct conditions).

\subsection{Evaluation Metrics}

We report four metrics for each model-task-condition combination. Following standard practice in binary classification evaluation, we use the F1 score as the primary metric for comparing conditions. F1 balances precision and recall, providing a single summary statistic that accounts for both false positives and false negatives. All statistical comparisons use paired t-tests over F1 scores across model-task combinations.

We additionally report:
\begin{itemize}
    \item \textbf{Accuracy}: The proportion of correct classifications, providing an intuitive measure of overall performance.
    \item \textbf{Precision}: The proportion of positive predictions that are correct, relevant for applications where false positives carry high costs.
    \item \textbf{Recall}: The proportion of actual positives correctly identified, relevant for applications where false negatives carry high costs (e.g., compliance screening).
\end{itemize}

We assess the significance of performance differences using paired t-tests over the 33 model-task combinations (11 models $\times$ 3 tasks). Effect sizes are reported as Cohen's $d$. We consider $p < 0.05$ as the threshold for statistical significance.

Results are aggregated at three levels: (1) overall averages across all model-task combinations, (2) task-level averages across models, and (3) model-level averages across tasks. This multi-level analysis enables identification of both general patterns and specific model-task interactions.

\subsection{Implementation and Reproducibility}

Task definitions are formalised as OWL 2 ontologies with SWRL rules encoding classification logic. We use the owlready2 Python library (v0.45) for ontology manipulation and the Pellet reasoner~\cite{Sirin2007} for SWRL rule execution. During evaluation, a fresh OWL world is instantiated for each test instance to ensure isolation between examples. Extracted entities and assertions are translated to ABox assertions, the reasoner performs inference, and class membership of the target individual determines the classification outcome.

All LLM interactions use the LangChain framework (v0.3.x) with provider-specific API clients. Structured output is enforced via Pydantic models specifying the expected response schema for entity identification and assertion extraction steps.

Experiments were conducted using Python 3.13 on Linux systems. API calls were made to provider endpoints (OpenAI, Anthropic, Google Gemini API, Fireworks AI) between 2025-12-01 and 2025-12-20.

We provide the following materials to support reproducibility:
\begin{itemize}
    \item Dataset generation code: \url{https://github.com/albsadowski/ruleeval-xd}
    \item Framework implementation including ontologies, prompts, and evaluation scripts: \url{https://github.com/albsadowski/structured-decomposition-swj}
\end{itemize}

\noindent The OWL ontologies for all three tasks, including TBox definitions, SWRL rules, and \texttt{sd:} namespace annotations for LLM prompting, are included in the framework repository.

\section{Results}
\label{section:results}

We evaluated structured decomposition (SD) against baselines across 11 models and 3 tasks, yielding 33 model-task combinations. The experimental design includes four SD variants to isolate the contributions of symbolic verification and complementary predicates. Table~\ref{tab:main_results} summarises performance across all conditions; full results for each model-task combination are provided in Table~\ref{tab:detailed_results}, with ablation comparisons in Table~\ref{tab:ablation_results}.

\subsection{Overall Performance}

\begin{figure}
\begin{center}
\includegraphics[width=0.5\columnwidth]{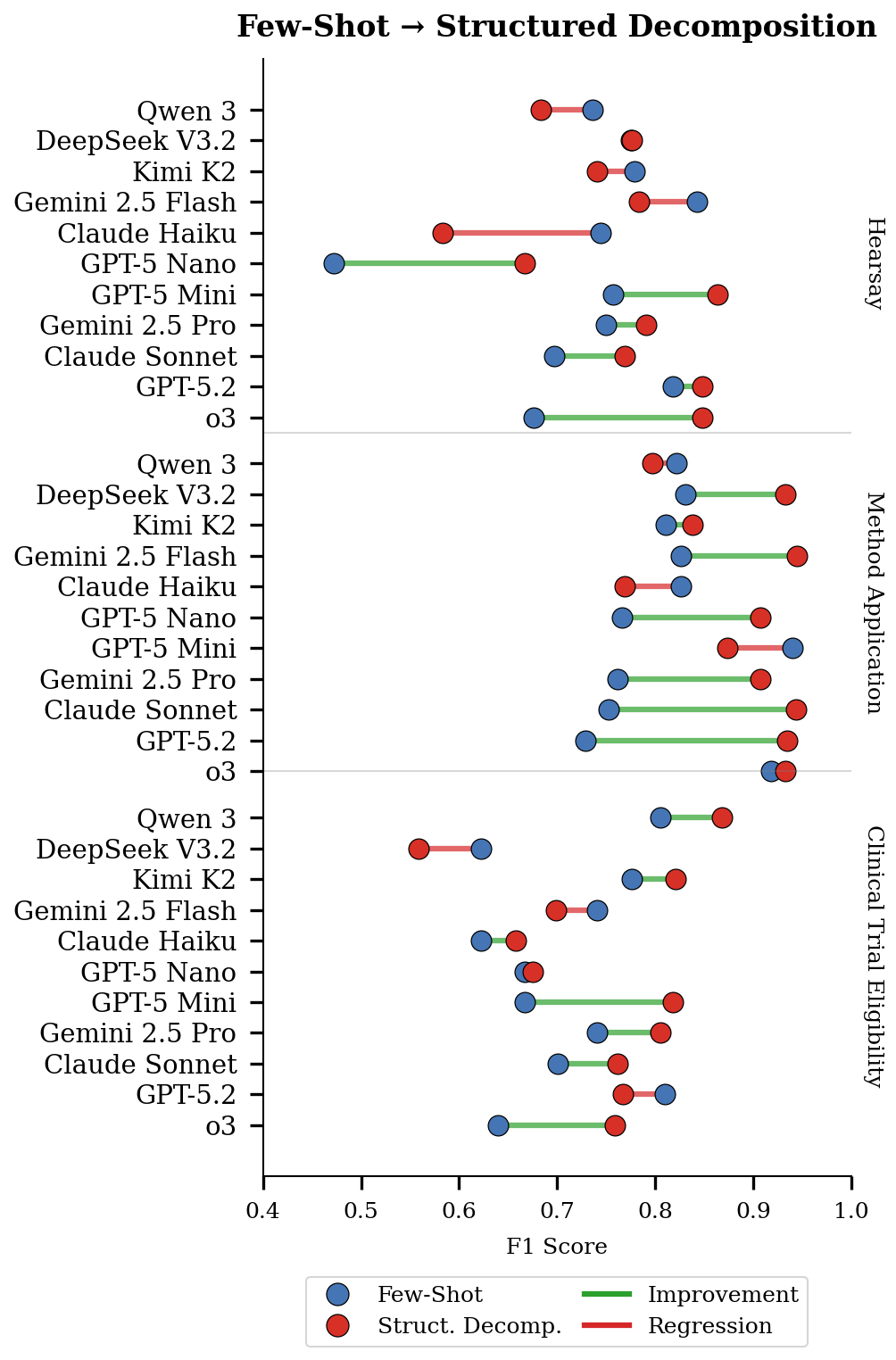}
\end{center}
\caption{Performance shift from few-shot baseline (blue) to structured decomposition (red) across eleven models and three tasks. Green lines indicate improvement; red lines indicate regression. Method Application shows the most consistent gains.}
\label{fig:improvement}
\end{figure}

Structured decomposition without complementary predicates achieved the highest average F1 score (79.8\%) compared to few-shot prompting (75.2\%), chain-of-thought (74.1\%), and the SD-Direct ablation (70.1\%). The improvement over few-shot prompting was 4.6 percentage points and statistically significant according to a paired t-test ($t(32) = 2.88$, $p = 0.007$, Cohen's $d = 0.5$). The improvement over chain-of-thought was 5.7 percentage points ($t(32) = 2.84$, $p = 0.008$, $d = 0.5$). Figure~\ref{fig:improvement} illustrates these performance shifts across all model-task combinations.

As shown in Figure~\ref{fig:improvement}, the Method Application task exhibits the most consistent improvements, with nearly all models benefiting from structured decomposition. The Hearsay and Clinical Trial Eligibility tasks show greater variability, with some models experiencing regression, suggesting that framework effectiveness depends on both task characteristics and model.

\begin{table}[t]
\centering
\caption{Average performance across all 33 model-task combinations. SD achieves the highest F1 while maintaining notably higher recall than baseline methods. SD-C = SD with complementary predicates.}
\label{tab:main_results}
\begin{tabularx}{\columnwidth}{lXXXX}
\toprule
\textbf{Method} & \textbf{Acc.} & \textbf{Prec.} & \textbf{Recall} & \textbf{F1} \\
\midrule
Few-shot & 78.4\% & 85.3\% & 69.0\% & 75.2\% \\
CoT & 77.6\% & 85.7\% & 68.2\% & 74.1\% \\
SD & 80.0\% & 79.4\% & 84.7\% & 79.8\% \\
SD-C & 77.4\% & 81.2\% & 74.2\% & 74.8\% \\
SD-Direct & 73.6\% & 77.7\% & 71.2\% & 70.1\% \\
SD-Direct-C & 76.4\% & 79.8\% & 73.4\% & 72.5\% \\
\bottomrule
\end{tabularx}
\end{table}

\subsection{Ablation: Contribution of Symbolic Verification}

To isolate the contribution of the symbolic reasoning component, we compare SD against SD-Direct, an ablation condition that uses identical assertion specifications and extraction prompts but bypasses the OWL/SWRL verification step, instead asking the LLM to directly determine the classification from extracted predicates. SD outperformed SD-Direct by 9.7 percentage points ($t(32) = 3.71$, $p = 0.001$, $d = 0.65$), confirming that the symbolic verification component provides substantial benefit beyond the structured predicate definitions alone. Table~\ref{tab:ablation_results} presents the complete ablation results.

\begin{table*}[t]
\centering
\caption{Ablation study: F1 scores (\%) comparing structured decomposition with SWRL verification (SD, SD-C) against variants without symbolic reasoning (SD-D, SD-D-C). Best result per row within each task is shown in \textbf{bold}.}
\label{tab:ablation_results}
\small
\begin{tabular}{l cccc cccc cccc}
\toprule
& \multicolumn{4}{c}{\textbf{Hearsay}} & \multicolumn{4}{c}{\textbf{Method Application}} & \multicolumn{4}{c}{\textbf{Clinical Trial Eligibility}} \\
\cmidrule(lr){2-5} \cmidrule(lr){6-9} \cmidrule(lr){10-13}
\textbf{Model} & SD & SD-C & SD-D & SD-D-C & SD & SD-C & SD-D & SD-D-C & SD & SD-C & SD-D & SD-D-C \\
\midrule
o3 & 84.8 & 79.1 & \textbf{88.6} & 85.1 & 93.3 & \textbf{95.2} & 92.2 & 95.1 & \textbf{75.9} & 66.7 & 62.2 & 71.4 \\
GPT-5.2 & 84.8 & 84.7 & 77.6 & \textbf{89.7} & 93.5 & \textbf{94.3} & 85.2 & 94.2 & \textbf{76.7} & 50.7 & 60.9 & 56.3 \\
GPT-5 Mini & \textbf{86.4} & 77.1 & 82.2 & 84.4 & 87.4 & 82.1 & \textbf{91.7} & 90.7 & \textbf{81.8} & 70.0 & 62.2 & 63.0 \\
GPT-5 Nano & 66.7 & 63.5 & \textbf{72.4} & 70.6 & \textbf{90.7} & 80.8 & 88.5 & 87.9 & \textbf{67.5} & 64.0 & 58.3 & 59.2 \\
Claude 4.5 Sonnet & 76.9 & 76.1 & \textbf{79.5} & 78.5 & \textbf{94.4} & 87.8 & 49.3 & 77.6 & \textbf{76.2} & 55.1 & 36.6 & 45.9 \\
Claude 4.5 Haiku & 58.3 & 72.9 & 69.0 & \textbf{71.3} & 76.9 & 81.4 & 79.2 & \textbf{81.5} & \textbf{65.8} & 62.2 & 47.1 & 41.3 \\
Gemini 2.5 Pro & 79.1 & 72.7 & \textbf{87.4} & 84.4 & \textbf{90.7} & 91.6 & 76.2 & 84.7 & \textbf{80.5} & 71.8 & 70.1 & 69.6 \\
Gemini 2.5 Flash & \textbf{78.4} & 78.0 & 72.2 & 72.1 & \textbf{94.5} & 93.7 & 92.6 & 91.7 & \textbf{69.9} & 66.7 & 60.4 & 64.4 \\
DeepSeek v3.2 & \textbf{77.6} & 61.5 & 48.1 & 29.2 & \textbf{93.3} & 91.7 & 65.8 & 90.2 & \textbf{55.9} & 37.3 & 8.2 & 8.2 \\
Kimi K2 & 74.1 & 77.2 & 77.6 & \textbf{79.1} & \textbf{83.8} & 78.3 & 81.5 & 82.5 & 82.1 & \textbf{83.0} & 71.4 & 73.2 \\
Qwen 3 & 68.3 & 62.9 & 73.1 & \textbf{73.3} & 79.7 & 78.7 & 82.2 & \textbf{82.9} & \textbf{86.8} & 78.0 & 63.0 & 64.9 \\
\midrule
\textbf{Average} & \textbf{75.9} & 73.2 & 75.2 & 74.3 & \textbf{88.9} & 86.9 & 80.4 & 87.1 & \textbf{74.4} & 64.1 & 54.5 & 56.1 \\
\bottomrule
\end{tabular}
\end{table*}

This ablation result is notable: providing structured assertion specifications without symbolic verification actually \textit{degrades} performance compared to few-shot prompting (70.1\% vs.\ 75.2\% F1). The decomposition into predicates appears to introduce additional opportunities for error that only pay off when coupled with formal verification. This finding underscores that architectural decomposition with symbolic reasoning drives the observed improvements.

The contribution of symbolic verification varies by task. On Clinical Trial Eligibility, the difference was particularly pronounced: SD achieved 74.5\% F1 compared to 54.6\% for SD-Direct, a gap of 19.9 percentage points. Method Application showed a more modest but still substantial difference of 8.5 percentage points (88.9\% vs.\ 80.4\%). On Hearsay, the difference was minimal (75.9\% vs.\ 75.2\%), suggesting that for tasks requiring nuanced interpretation, the extraction step dominates overall performance regardless of verification method.

\subsection{Ablation: Effect of Complementary Predicates}

Contrary to findings from our prior work~\cite{sadowski2025explainable}, complementary predicates significantly degraded performance in the current evaluation. SD without complementary predicates outperformed SD with complementary predicates (SD-C) by 5.0 percentage points (79.8\% vs.\ 74.8\%, $t(32) = 3.69$, $p = 0.001$, $d = 0.64$).

This effect was most pronounced on Clinical Trial Eligibility, where complementary predicates reduced F1 by 10.3 percentage points (74.5\% vs.\ 64.1\%). On Hearsay, the reduction was 2.8 percentage points (75.9\% vs.\ 73.2\%), and on Method Application, 2.1 percentage points (88.9\% vs.\ 86.9\%).

The pattern persisted in the SD-Direct conditions: SD-Direct without complementary predicates outperformed the variant with complementary predicates on Hearsay (75.2\% vs.\ 74.3\%) and Clinical Trial Eligibility (54.5\% vs.\ 56.1\%), though the relationship was mixed on Method Application (80.4\% vs.\ 87.1\%).

\subsection{Precision-Recall Characteristics}

\begin{figure*}
\setlength{\fboxsep}{0pt}%
\setlength{\fboxrule}{0pt}%
\begin{center}
\includegraphics{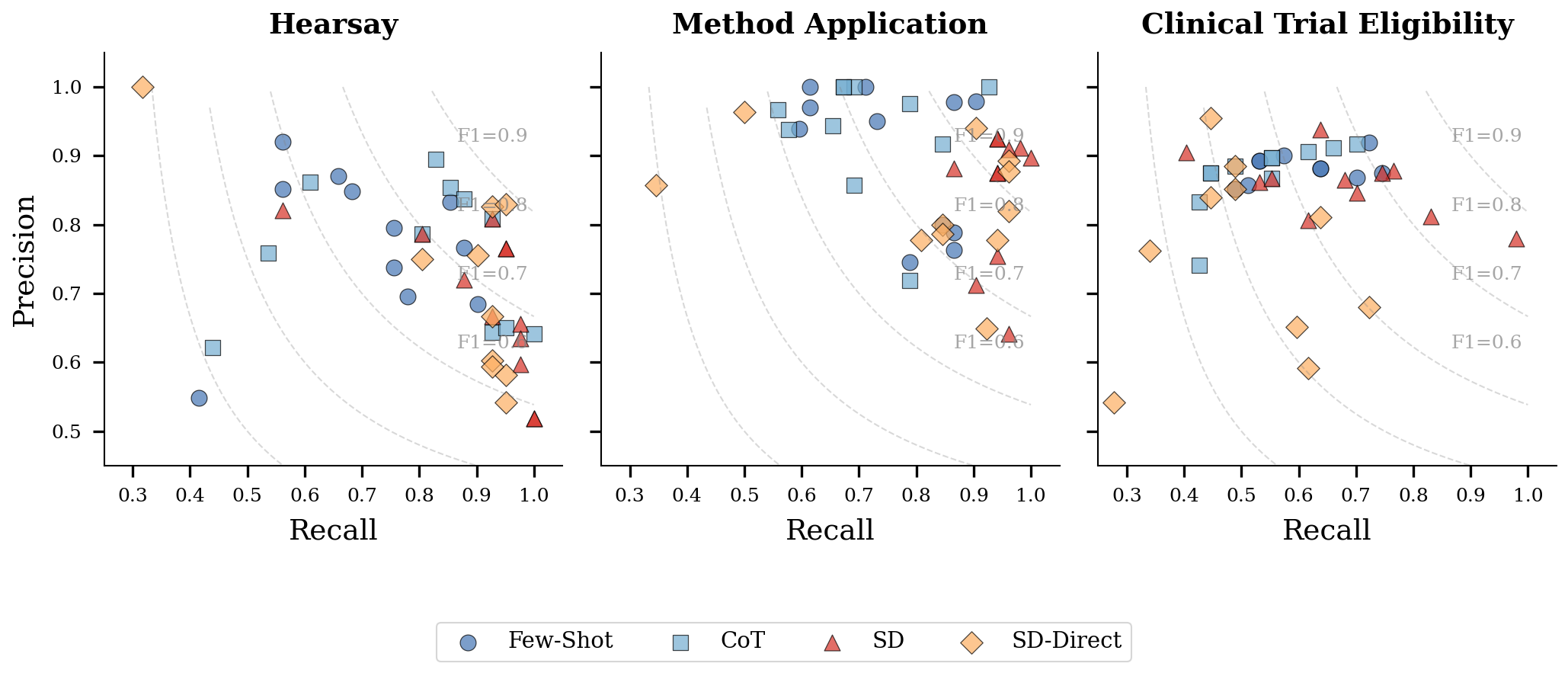}
\end{center}
\caption{Precision-recall trade-offs by prompting strategy. Dashed curves show F1 iso-lines (0.6--0.9). Structured decomposition (triangles) achieves higher recall, with strongest performance on Method Application, supporting generalisation beyond the legal domain.}
\label{fig:precision_recall}
\end{figure*}

A notable pattern emerged in the precision-recall trade-offs (Figure~\ref{fig:precision_recall}). SD achieved substantially higher recall (84.7\%) compared to few-shot (69.0\%) and chain-of-thought (68.2\%). This difference of 15.7 percentage points over few-shot was highly significant ($t(32) = 7.00$, $p < 0.001$, $d = 1.22$). The precision trade-off was modest: SD precision (79.4\%) was lower than few-shot (85.3\%), but the recall gains more than compensated in terms of overall F1, as evidenced by SD points clustering toward higher iso-lines in Figure~\ref{fig:precision_recall}.

\subsection{Task-Level Analysis}

Table~\ref{tab:task_results} shows the breakdown by task, including the proportion of models for which SD outperformed few-shot prompting.

\begin{table}[t]
\centering
\caption{SD performance by task. SD $>$ FS indicates proportion of models where SD outperformed few-shot prompting.}
\label{tab:task_results}
\begin{tabular}{lccc}
\toprule
\textbf{Task} & \textbf{$\Delta$ F1} & \textbf{$p$} & \textbf{SD $>$ FS} \\
\midrule
Hearsay & +2.8pp & .402 & 7/11 \\
Method Application & +7.3pp & .034 & 8/11 \\
Clinical Trial Eligibility & +3.6pp & .104 & 8/11 \\
\bottomrule
\end{tabular}
\end{table}

SD performed best on the \textit{method application} task, achieving a 7.3 percentage point improvement over few-shot ($p = 0.008$) with 8 out of 11 models showing gains.

On \textit{clinical trial eligibility}, SD outperformed few-shot on 8 out of 11 models with an aggregate improvement of 3.6 percentage points, though this did not reach statistical significance. Individual results revealed high variance: SD produced substantial gains on some models (e.g., Qwen 3 at +6.3pp, GPT-5 Mini at +15.1pp) while underperforming on others (e.g., GPT-5.2 at $-$4.3pp, DeepSeek v3.2 at $-$6.3pp).

The \textit{hearsay} task showed the most variability, with 7 out of 11 models improving and modest aggregate gains (+2.8pp). Some models showed large regressions (e.g., Claude 4.5 Haiku at $-$16.1pp, Qwen 3 at $-$5.3pp).

\subsection{Individual Model-Task Results}

The largest SD improvements over few-shot on \textit{method application} were GPT-5.2 (+20.6pp), Claude 4.5 Sonnet (+19.1pp), Gemini 2.5 Pro (+14.5pp), and GPT-5 Nano (+14.1pp). On \textit{clinical trial eligibility}: GPT-5 Mini (+15.1pp), o3 (+11.9pp), and Kimi K2 (+4.5pp). On \textit{hearsay}: GPT-5 Nano (+19.5pp), o3 (+17.2pp), and GPT-5 Mini (+10.7pp).

Conversely, the largest regressions appeared primarily on \textit{hearsay}: Claude 4.5 Haiku ($-$16.1pp), Gemini 2.5 Pro ($-$7.0pp), and Qwen 3 ($-$5.3pp).

\subsection{Model-Level Patterns}

Aggregating across tasks, nine of eleven models showed positive average improvement with SD. o3 showed the strongest gains (+9.5pp average), achieving the highest overall SD performance (84.7\% F1). GPT-5.2 and Claude 4.5 Sonnet showed similarly strong improvements (+6.3pp and +7.5pp respectively). The GPT family showed consistent positive gains across all variants.

Three models showed degradation: Claude 4.5 Haiku ($-$6.4pp), Qwen 3 ($-$2.9pp), and Gemini 2.5 Flash ($-$1.1pp). Within model families, larger variants tended to show greater improvement: Gemini 2.5 Pro (+6.9pp) versus Gemini 2.5 Flash ($-$1.1pp), and Claude 4.5 Sonnet (+7.5pp) versus Claude 4.5 Haiku ($-$6.4pp).

\begin{table*}[t]
\centering
\caption{F1 scores (\%) for all model-task combinations. Best result per row is shown in \textbf{bold}. FS = few-shot, CoT = chain-of-thought, SD = structured decomposition (without complementary predicates), SD-C = SD with complementary predicates.}
\label{tab:detailed_results}
\small
\begin{tabular}{l cccc cccc cccc}
\toprule
& \multicolumn{4}{c}{\textbf{Hearsay}} & \multicolumn{4}{c}{\textbf{Method Application}} & \multicolumn{4}{c}{\textbf{Clinical Trial Eligibility}} \\
\cmidrule(lr){2-5} \cmidrule(lr){6-9} \cmidrule(lr){10-13}
\textbf{Model} & FS & CoT & SD & SD-C & FS & CoT & SD & SD-C & FS & CoT & SD & SD-C \\
\midrule
o3 & 67.6 & 71.4 & \textbf{84.8} & 79.1 & 91.8 & 90.5 & 93.3 & \textbf{95.2} & 64.0 & 63.0 & \textbf{75.9} & 66.7 \\
GPT-5.2 & 81.8 & 77.2 & \textbf{84.8} & 84.7 & 72.9 & 70.7 & 93.5 & \textbf{94.3} & \textbf{81.0} & 76.5 & 76.7 & 50.7 \\
GPT-5 Mini & 75.7 & 62.9 & \textbf{86.4} & 77.1 & \textbf{94.0} & 87.2 & 87.4 & 82.1 & 66.7 & 67.5 & \textbf{81.8} & 70.0 \\
GPT-5 Nano & 47.2 & 51.4 & \textbf{66.7} & 63.5 & 76.6 & 75.2 & \textbf{90.7} & 80.8 & \textbf{66.7} & 59.2 & 67.5 & 64.0 \\
Claude 4.5 Sonnet & 69.7 & 76.0 & \textbf{76.9} & 76.1 & 75.3 & 71.4 & \textbf{94.4} & 87.8 & 70.1 & 54.1 & \textbf{76.2} & 55.1 \\
Claude 4.5 Haiku & 74.4 & \textbf{78.1} & 58.3 & 72.9 & 82.6 & \textbf{88.0} & 76.9 & 81.4 & 62.2 & 56.3 & \textbf{65.8} & 62.2 \\
Gemini 2.5 Pro & 75.0 & \textbf{86.1} & 79.1 & 72.7 & 76.2 & 80.5 & \textbf{90.7} & 91.6 & 74.1 & 79.5 & \textbf{80.5} & 71.8 \\
Gemini 2.5 Flash & 84.3 & \textbf{85.4} & 78.4 & 78.0 & 82.6 & 81.8 & \textbf{94.5} & 93.7 & \textbf{74.1} & 68.4 & 69.9 & 66.7 \\
DeepSeek v3.2 & 77.5 & \textbf{79.5} & 77.6 & 61.5 & 83.1 & 80.5 & \textbf{93.3} & 91.7 & \textbf{62.2} & 59.2 & 55.9 & 37.3 \\
Kimi K2 & 77.9 & \textbf{86.4} & 74.1 & 77.2 & 81.1 & 77.3 & \textbf{83.8} & 78.3 & 77.6 & 68.4 & 82.1 & \textbf{83.0} \\
Qwen 3 & 73.6 & \textbf{85.7} & 68.3 & 62.9 & \textbf{82.2} & 76.6 & 79.7 & 78.7 & 80.5 & 73.4 & \textbf{86.8} & 78.0 \\
\midrule
\textbf{Average} & 73.2 & 76.4 & \textbf{75.9} & 73.2 & 81.7 & 80.0 & \textbf{88.9} & 86.9 & 70.8 & 66.0 & \textbf{74.4} & 64.1 \\
\bottomrule
\end{tabular}
\end{table*}

\section{Discussion}
\label{section:discussion}

Across all 33 model-task combinations, structured decomposition achieved an average F1 of 79.8\%, compared to 75.2\% for few-shot and 74.1\% for chain-of-thought. The improvement of 4.6 percentage points was statistically significant ($t(32) = 2.88$, $p = 0.007$, Cohen's $d = 0.50$), with nine of eleven models showing positive gains. Method application showed the largest gains (+7.3pp), followed by clinical trial eligibility (+3.6pp) and hearsay (+2.8pp), confirming that the framework generalises beyond the legal domain where it was originally validated.

However, generalisability is bounded by task suitability. As demonstrated by the negative validation on URTI classification (Section~\ref{subsec:task-selection}), structured decomposition requires two properties: a rule-expressible decision boundary and formalisable predicate structure. Tasks where identical observable features can correspond to different labels, such as medical diagnosis from symptom profiles, require statistical discrimination that logical rules cannot provide. The framework is not a general-purpose improvement but a targeted approach for rule-governed domains.

\subsection{The Contribution of Symbolic Verification}

The ablation study establishes that SWRL-based reasoning provides substantial benefit beyond structured prompting alone. SD outperformed SD-Direct by 9.7 percentage points overall, with the gap reaching 19.9 percentage points on clinical trial eligibility. This difference confirms that the symbolic reasoner contributes meaningfully to classification accuracy rather than serving as architectural overhead.

A notable finding is that structured prompting without symbolic verification actually degrades performance compared to few-shot baselines (70.1\% vs.\ 75.2\% F1). Decomposing the task into entity identification and assertion extraction introduces additional points where errors can occur. These errors compound when the LLM must also perform the final classification. Symbolic verification appears to correct or constrain these errors, yielding net improvement. Without it, the added complexity provides no benefit.

The contribution of symbolic verification varied by task. On clinical trial eligibility, where the reasoning involves matching patient characteristics against explicit criteria, the reasoner's contribution was largest. On hearsay, where predicate extraction requires interpreting legal language with considerable ambiguity, the gap between SD and SD-Direct was minimal. This suggests that when extraction difficulty dominates, verification cannot compensate for upstream errors.

\subsection{Rule Complexity and Framework Benefits}

The task predicates in this evaluation are relatively simple logical structures. Hearsay determination is a conjunction of five conditions; method application requires four predicates; clinical trial eligibility uses three. All are straightforward conjunctions without disjunctions, nested quantifiers, or complex logical patterns. One might expect that such simple rules would not benefit substantially from formal verification, as LLMs should be capable of evaluating a handful of conjuncts directly.

Yet the results show otherwise. Method application, with only four predicates, showed the largest improvement over few-shot (+7.3pp) and a substantial gap between SD and SD-Direct (88.9\% vs.\ 80.4\%). Clinical trial eligibility, with only three predicates, showed the widest gap between SD and SD-Direct: 19.9 percentage points (74.5\% vs.\ 54.6\%). The simplicity of the rules did not diminish the framework's contribution.

This finding bears on the role of SD-Direct. One rationale for allowing the LLM to determine the final classification from extracted predicates is that it might reconcile extraction deficiencies: seeing all predicates together, the model could correct inconsistencies or adjust for low-confidence extractions. The results suggest this does not occur. SD-Direct consistently underperformed SD on method application and clinical trial eligibility. Rather than correcting errors, the additional inference step appears to introduce new ones.

The implication is that even simple logical rules benefit from external verification. The difficulty lies not in the complexity of the rule but in the accumulation of small extraction errors across predicates. Each predicate extraction carries some probability of error; when the LLM must also perform the final logical combination, these errors compound. Symbolic verification eliminates one source of error entirely, yielding improvements even when the underlying logic is trivial.

\subsection{Complementary Predicates and Ontology Design}

Contrary to expectations from our prior work~\cite{sadowski2025explainable}, complementary predicates degraded performance by 5.0 percentage points overall ($p = 0.001$), with the largest effect on clinical trial eligibility ($-$10.3pp). Our prior work hypothesised that complementary predicates would improve precision by forcing explicit consideration of both positive and negative conditions, addressing confirmation bias and grounding reasoning under OWL's open-world assumption. However, even in that evaluation, benefits were concentrated in specific models (o1: +14.5pp F1, o3-mini: +13.7pp) while others showed mixed effects (Claude 3.7 Sonnet: $-$1.3pp). The current evaluation, using a different and more recent model set, observes consistently negative effects across GPT-5 variants, Claude 4.5, and Gemini 2.5. Two factors likely contribute: the benefit appears highly model-specific rather than universal, and confirmation bias may be less prevalent in current-generation LLMs, reducing the benefit while the additional cognitive load of evaluating paired assertions introduces extraction errors.

This finding has practical implications for ontology design in neural-symbolic systems. When LLMs handle entity and predicate extraction, simpler ontologies with fewer assertions appear preferable. The theoretical benefits of explicit negation under the open-world assumption do not outweigh the practical costs of increased extraction complexity. Practitioners designing task ontologies should favour minimal predicate sets that capture necessary conditions without redundant complementary assertions, and empirically validate any complementary predicates for their specific model deployments rather than assuming benefit.

\subsection{Precision-Recall Trade-offs}

SD achieved substantially higher recall than baselines (84.7\% vs.\ 69.0\% for few-shot), with a modest reduction in precision (79.4\% vs.\ 85.3\%). This pattern was consistent across all three tasks.

The recall advantage suggests that structured decomposition may be particularly suited to applications where false negatives carry higher costs than false positives. Compliance screening, eligibility determination, and evidence review are examples where missing a relevant case is more costly than flagging a borderline one for human review. The framework's tendency toward inclusive classification aligns with these use cases.

For applications where false positives are more costly, the framework's precision-recall profile may be less appropriate, or additional filtering mechanisms may be needed downstream.

\subsection{Model Capability Requirements}

Within model families, larger and more capable variants showed greater improvement from structured decomposition. Gemini 2.5 Pro gained 8.3 percentage points while Gemini 2.5 Flash 0.6 percentage points. Claude 4.5 Sonnet gained 10.8 percentage points while Claude 4.5 Haiku lost 6.1 percentage points.

This pattern reflects the framework's architecture. The decomposition approach relies on the model to correctly identify entities and extract assertions according to ontology specifications. Models with stronger instruction-following and reasoning capabilities perform this extraction more accurately. The symbolic verification step guarantees correct rule application given the extracted predicates, but it cannot compensate for extraction errors in earlier steps. When extraction quality is poor, the framework's benefits do not materialise.

Practitioners should therefore consider model capability when deciding whether to deploy structured decomposition. The framework is not a substitute for model quality but rather a way to leverage capable models more effectively on rule-governed tasks.

\subsection{Benefits of Semantic Web Integration}

The \ref{section:introduction} motivated our transition from SMT to OWL/SWRL on pragmatic grounds (reliability at scale) while noting strategic advantages in inference expressivity. Here we assess these benefits in light of the experimental results.

The populated ABox provides a complete reasoning trace. Each entity extracted, each assertion evaluated, and each justification persists as an explicit record. When a classification is incorrect, practitioners can inspect the ABox to identify which extraction step failed. This traceability derives from the decomposition architecture itself rather than from OWL specifically; the choice of representation affects how traces are stored and queried, not whether they exist. However, the ABox integrates with standard semantic web tooling without requiring separate infrastructure for rule execution and result storage. The SPARQL query in Section~\ref{section:framework} demonstrates that extraction decisions are queryable directly. We do not claim SPARQL provides capabilities unavailable in relational databases; the value lies in a unified formalism spanning task definition, rule evaluation, and result inspection.

Task definitions become reusable artifacts editable in standard ontology tooling such as Protégé and validatable for logical consistency prior to deployment. The separation between task specification (TBox) and instance data (populated ABox) supports iterative refinement by domain experts without modifying framework code.

The OWL foundation positions the framework for inference patterns beyond those evaluated here. Class hierarchies could enable structured multi-class output where predictions belong to multiple abstraction levels. Multiple SWRL rules firing independently over the same ABox could support multi-label classification without combinatorial rule explosion. Composable ontologies could import established domain vocabularies such as SNOMED-CT for medical applications or FIBO for financial domains. Our evaluation focused on binary classification tasks, though the architectural foundation supports these richer patterns without modification.

\subsection{Limitations and Future Work}

Several limitations constrain the interpretation of these results and suggest directions for future work.

All three evaluation tasks involve binary classification. The framework's applicability to multi-class classification, ranking, or extraction tasks remains untested. Similarly, the task predicates evaluated here are simple conjunctions; whether benefits extend to more complex logical structures involving disjunctions, negation, or nested quantifiers is unknown.

Ontology design required manual effort by the authors. Creating effective entity and assertion descriptions demands domain understanding and iterative refinement. The effort required for new domains was not systematically measured. Recent work on multi-agent frameworks for automated knowledge acquisition~\cite{sadowski2025solar} suggests a potential path toward reducing this burden, though such approaches introduce their own validation challenges.

We evaluated only English-language tasks; performance on other languages, where LLM capabilities may differ, is unknown.

Finally, the interaction between model capability and framework effectiveness suggests that as LLMs continue to improve, the optimal division of labour between neural extraction and symbolic verification may shift. Longitudinal evaluation across model generations would help characterise this relationship.

\section{Conclusion}
\label{section:conclusion}

This paper presented an integration pattern that combines LLMs with OWL/SWRL reasoning for rule-based tasks over natural language input. The core architectural insight is that LLMs can serve as ontology population engines: components that translate unstructured text into ABox assertions according to expert-authored TBox specifications. This division exploits the complementary strengths of neural and symbolic approaches. LLMs handle the interpretive work of understanding natural language, while SWRL-based reasoners apply rules with deterministic guarantees. The populated ABox integrates with standard semantic web tooling for inspection and querying, and the formulation positions the framework for inference patterns that simpler formalisms cannot express: class hierarchies, multi-label classification, and composable imports from established domain vocabularies.

The paper makes two contributions. First, we demonstrate that grounding neural-symbolic integration in semantic web standards yields a principled architecture for rule-based reasoning. The separation between expert-authored TBox specifications and LLM-populated ABox assertions creates inspectable reasoning traces where every extraction decision persists as a queryable record. Second, we validate this pattern empirically across three domains and a diverse set of language models. Structured decomposition achieved statistically significant improvements over few-shot prompting in aggregate, with gains observed across legal reasoning, scientific text analysis, and medical inference. An ablation study confirmed that symbolic verification provides substantial benefit beyond structured prompting alone: removing the SWRL reasoning step markedly degraded performance.

The pattern offers a principled answer to a recurring question in neural-symbolic integration: how should LLMs and formal reasoning systems be combined? By grounding the integration in semantic web standards, the approach inherits decades of tooling, well-defined semantics, and established practices for ontology engineering. The result is a framework where LLMs do what they do best, interpret language, while formal systems provide the guarantees that high-stakes domains require.

\bibliographystyle{unsrt}

\end{document}